\documentclass[11pt]{article}

\usepackage[final]{acl}

\usepackage{times}
\usepackage{latexsym}

\usepackage[T1]{fontenc}
\usepackage{amsmath}
\usepackage{amssymb}
\usepackage{booktabs}
\usepackage{wrapfig}
\usepackage{float}
\usepackage{multirow}

\usepackage[utf8]{inputenc}

\usepackage{microtype}

\usepackage{inconsolata}

\usepackage{graphicx}

\title{KV-Skill: Forging Expertise in the Model's Native Language}

\author{
Zhaowei Han\thanks{Equal contribution.} \quad
Xiang Zhang\footnotemark[1] \quad
Bing Han \quad
Kai Liu \quad
Danqi Hu \quad
Jie Liu \\
University of Michigan, Ann Arbor, MI, USA \\
\texttt{rickyhan@umich.edu}
}

\begin{document}
\maketitle
\begin{abstract}
Task knowledge is commonly stored either as text in the prompt or as an update
to model weights. Text is modular but must be interpreted on every use, while
weight adaptation makes the resulting capability difficult to load, remove, or
share independently. We introduce \textbf{KV-Skill}, a design space of external
factorized operators that a frozen language model reads through a lightweight
interface. KV-Skill supports two complementary paths. Registration converts an
authored text skill into a text-derived operator and trains a shared
per-backbone interface. Reward learning develops a compact latent operator
directly from task outcomes, with or without an authored skill. Neither path
adds positions to the prompt. Across ten benchmarks and four backbones from
three model families, converting text to a KV-Skill consistently makes the same procedural
knowledge more effective on mathematical reasoning and retrieval tasks. On Qwen3.5-4B LiveMath, registration reaches 77.2
accuracy, compared with 52.0 for the same skill as text (SkillOpt) and 64.5 for SoftSkill. Under matched reward training and parameter budgets,
KV-Skill gives the best result in seven of eight matched settings against soft
prefixes, prefix tuning, and LoRA. A post-hoc rank analysis further shows that text-derived operators
retain nearly all of their benefit with one task-aligned direction per
injection layer, while matched random directions fail. Finally, a single shared interface serves \emph{eleven diverse skills} at once with
no loss in average accuracy. These results show that task knowledge can be acquired from text or
experience, compressed into an external operator, and deployed separately from
the backbone. Code is available at: \url{https://github.com/shawnzhg/KV-Skill}
\end{abstract}

\section{Introduction}

Task knowledge has a different lifecycle from general model knowledge. It may
begin as a written procedure, emerge from successful experience, or change as
a task evolves. It should be possible to acquire this knowledge once, store it
as an independent capability, and load it only when needed. Current approaches
couple these stages together. They make the form in which knowledge is acquired
also determine how it must be deployed.

Natural-language \emph{text skills} make the tradeoff clear
\citep{wang2024voyager,shinn2023reflexion,yang2026skillopt}. They are easy to
author, inspect, and share. Yet every use requires the model to read the skill
and translate its instructions into internal computation. A procedure can be
correct and still fail because the model cannot reliably execute it from text.
Post-training solves the access problem by learning directly in the model, but
stores the capability inside the weights. Continuous prompts and compressed
contexts remain external, although they still occupy the model's attention
state and are commonly trained through imitation
\citep{li2021prefix,lester2021power,eyuboglu2026cartridges,tao2026softskill}.
This leaves a gap between the modularity of text and the effectiveness of
model-native adaptation. We ask:

\begin{quote}
\emph{Can task knowledge be represented in a model-native form, loaded on demand, and developed directly from experience?}
\end{quote}

\begin{figure*}[t]
\centering
\includegraphics[width=\linewidth]{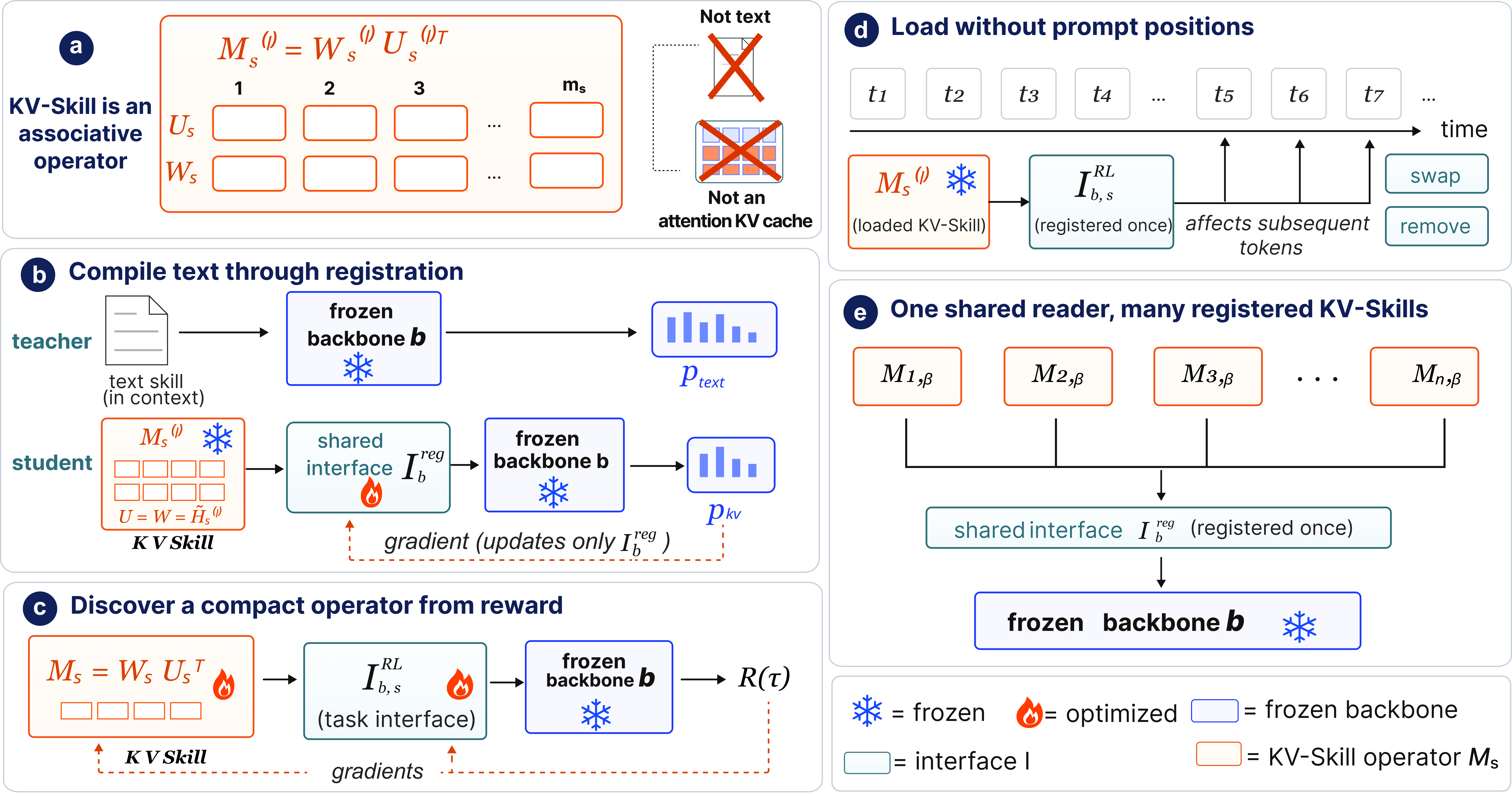}
\caption{\textbf{Overview of KV-Skill.}
(a) A KV-Skill is an external factorized associative operator, not text or an attention KV cache. Keys $U_s^{(\ell)}$ address slots; values $W_s^{(\ell)}$ return a query-dependent response.
(b) Registration converts a text skill into per-layer operators $M_s^{(\ell)}$, updating only the shared interface $I_b^{\mathrm{reg}}$.
(c) Reward learning optimizes a compact operator and task interface jointly, backbone frozen. The operator can start from scratch or from text-informed initialization.
(d) Loading, replacing, or removing a KV-Skill adds no prompt positions; a swap requires one re-prefill.
(e) One registered interface lets a frozen backbone read multiple independent KV-Skills, each constructed per backbone.}
\label{fig:KVskill_overview}
\end{figure*}

We introduce \textbf{KV-Skill}, a design space of external factorized
associative operators for frozen language models
(Figure~\ref{fig:KVskill_overview}). At an injection layer, the operator
$M_s=W_sU_s^{\top}$ maps a query from the current residual state to a
task-specific response. This read uses a separate residual branch rather than
the model's attention KV cache. It adds no prompt positions and does not grow
with the conversation. A lightweight interface $I$ connects the operator to a
backbone. The capability remains in $M_s$, so it can be stored, selected, and
replaced independently of the model weights.

The operator abstraction separates knowledge acquisition from deployment.
When an authored text skill is available, one prefill harvests a
high-resolution, text-derived operator. Registration then teaches a shared
per-backbone interface to access this fixed $M_s$. When no procedure is
available, reward learning develops a compact latent operator and its task
interface directly from verifier outcomes. It can also begin from an available
skill. These are not identical tensor layouts. They are two endpoints of one
design space: a detailed operator compiled from text and a compact operator
discovered from experience.

The experiments show why both endpoints matter. Converting a text skill
produces large gains without changing its derived operator. On Qwen3.5-4B
LiveMath, registration reaches 77.2 accuracy, compared with 52.0 for the same skill as text (SkillOpt) and 64.5 for SoftSkill. The advantage also holds
on gpt-oss-20b and structured knowledge retrieval. Swapping the loaded operator
while holding $I$ fixed removes most of the gain, confirming that task identity
resides in $M_s$ rather than being hidden in the interface.

Reward learning provides the complementary result. A compact KV-Skill can be
learned without any authored procedure. Under the same reward objective,
training budget, and parameter budget, it gives the best result in seven of
eight matched settings with soft prefixes, prefix tuning, and LoRA. The result is therefore not explained
by reward optimization alone. The representation through which reward acts
also matters.

The most revealing result concerns what registration extracts from text. After
registration, the token-level operator can be reduced to one task-aligned
direction per injection layer with almost no loss on LiveMath, SearchQA, or
STaRK-Prime. Matched random directions fail, showing that this is compression
of task-specific structure rather than a scale effect. A single shared interface also serves eleven separately loadable KV-Skills at
once with no loss in average accuracy. The
complete retention and replay results are reported in
Appendix~\ref{sec:app-retention}. Taken together, the results show that task
knowledge can be acquired from text or experience, compressed after
acquisition, and deployed as an external capability.

Our contributions are as follows:

\begin{itemize}
\item \textbf{External operator design space.}
We define KV-Skill as an external factorized associative operator, rather than
an attention cache or one fixed tensor layout. The operator is loaded through a
lightweight interface and adds no prompt positions.

\item \textbf{Two paths to an external capability.}
Registration compiles an authored text skill into a fixed operator. Reward
learning develops a compact operator from task outcomes, with or without an
authored skill, while keeping the backbone frozen.

\item \textbf{Controlled evaluation.}
Across ten benchmarks and four backbones, KV-Skills improve over text
and continuous-skill baselines. Matched reward experiments isolate the effect
of the trainable substrate from the effect of the objective.

\item \textbf{Mechanistic and modularity evidence.}
Rank and direction controls show that registration distills text into compact,
task-aligned conditional steering. Sequential registration shows that one
interface can retain several independently loadable KV-Skills.
\end{itemize}

\section{Related Work}
\label{sec:related}

\paragraph{Text skills and continuous prompts.}
LLM agents often store reusable procedures as text
\citep{wang2024voyager,shinn2023reflexion}. Trace2Skill extracts such
procedures from trajectories, while TextGrad, GEPA, and SkillOpt optimize them
using task feedback
\citep{ni2026trace2skill,yuksekgonul2024textgrad,
agrawal2026gepa,yang2026skillopt}. Text remains readable and portable, but must
be processed on every use. Prompt tuning, prefix tuning, and context compressors
instead encode task information in continuous attention states
\citep{lester2021power,li2021prefix,liu2022ptuning,
mu2023learning,chevalier2023autocompressors,eyuboglu2026cartridges}.
SoftSkill similarly converts a text skill into an imitation-trained prefix
\citep{tao2026softskill}. KV-Skill differs by using a separate residual branch.
It adds neither prompt positions nor entries to the attention KV cache and can
be optimized directly from verifier reward.

\paragraph{Weight adaptation and activation steering.}
Adapters and LoRA learn task behavior through model-specific parameter updates
\citep{hu2022lora}, while Text-to-LoRA predicts such updates from task
descriptions \citep{charakorn2025texttolora}. KV-Skill instead keeps the capability
in an external operator, with several registered KV-Skills sharing one
per-backbone interface. Activation steering, task vectors, and function vectors
modify behavior through directions in the residual stream
\citep{turner2023steering,hendel2023context,todd2024function,
belitsky2025kv}. Recent work also trains per-layer steering vectors with
reinforcement learning \citep{sinii2025bias,sinii2025small}. A post-hoc SVD can compress a registered KV-Skill to one direction per
injection layer (Section~\ref{sec:exp-rank}). This form looks like a steering
vector, but its strength depends on the current query. With a fixed
coefficient, it loses 25 points on LiveMath, and RL-trained bias-only steering
also trails KV-Skill (Table~\ref{tab:substrate}).

\paragraph{Associative operators and external memory.}
The factorization $M_s=W_sU_s^\top$ follows associative reads used in linear
attention and fast-weight memory
\citep{katharopoulos2020transformers,schlag2021linear,
yang2024parallelizing}; we do not claim this operator form as new. KBLaM uses
external key--value memory for declarative knowledge
\citep{wang2025kblam}, while Cache-to-Cache and Latent Cache Flow communicate
instance-specific information through model states
\citep{fu2026cachetocache,rossi2026latent}. KV-Skill instead uses the operator as a
persistent procedural capability that can be constructed from text or learned
from task reward.
\section{The KV-Skill Design Space}
\label{sec:method}

KV-Skill stores task knowledge outside both the prompt and backbone in an
external operator $M_s$. A lightweight interface $I$ reads this operator from
the residual stream of a frozen language model. The operator can be loaded,
replaced, or removed without changing backbone weights or consuming prompt
positions.

This abstraction supports two constructions. A \emph{text-derived KV-Skill}
converts an authored text skill into a fixed operator, whereas a
\emph{reward-learned KV-Skill} learns a compact operator directly from task
outcomes. They use different tensor layouts but share the same associative
read, representing two points in one design space: knowledge harvested from
text and knowledge discovered from experience. Table~\ref{tab:skill-design} in
the appendix summarizes their differences in source, representation, and
optimization.

\subsection{KV-Skill as an Associative Operator}
\label{sec:method-repr}

Let $\mathcal{D}$ be the set of layers where a KV-Skill is read. At each
injection layer $\ell\in\mathcal{D}$, the KV-Skill provides two factor
matrices:
\begin{equation}
U_s^{(\ell)}, W_s^{(\ell)}
\in
\mathbb{R}^{d_s\times m_s},
\end{equation}
where $m_s$ is the number of skill slots and $d_s$ is the skill-space
dimension. These factors define an implicit operator:
\begin{equation}
M_s^{(\ell)}
=
W_s^{(\ell)}U_s^{(\ell)\top}.
\label{eq:skill-operator}
\end{equation}
The full matrix is never constructed. The model applies the factorized form
directly:
\begin{equation}
\operatorname{Read}
\left(
M_s^{(\ell)},q_\ell
\right)
=
\frac{1}{\sqrt{m_s}}
W_s^{(\ell)}
\left(
U_s^{(\ell)\top}q_\ell
\right).
\label{eq:skill-read}
\end{equation}

The columns of $U_s^{(\ell)}$ act as keys: they measure the signed relevance of
each skill slot to the current query. The columns of $W_s^{(\ell)}$ act as
values: they return the corresponding response. The read is linear and does
not apply a softmax across slots.

We therefore use the name \emph{KV-Skill} for its associative key--value
structure. These skill factors are not entries in the model's attention KV
cache. They do not represent conversation tokens, do not use attention heads or
positional encoding, and do not grow with the context.

\paragraph{Formal definition.}
A KV-Skill is the factorized operator family
\begin{equation}
M_s
\triangleq
\left\{
\left(
U_s^{(\ell)},W_s^{(\ell)}
\right)
:
\ell\in\mathcal{D}
\right\},
\label{eq:kvskill-definition}
\end{equation}
together with any sharing of these factors across layers. This definition does
not require a fixed number of slots or one coordinate space. Text-derived
KV-Skills use separate factors at each depth, while reward-learned KV-Skills
share one compact factor pair across depths.

\subsection{Reading a KV-Skill}
\label{sec:method-read}

The interface $I$ connects the external operator to a frozen backbone. Given a
residual state $h_\ell$, it first constructs a query:
\begin{equation}
q_\ell
=
\operatorname{norm}
\left(
A\,\operatorname{RMSNorm}(h_\ell)
\right),
\label{eq:query}
\end{equation}
where $A$ maps the model state into the skill space. The interface then reads
the loaded operator using Equation~\ref{eq:skill-read} and writes the response
back:
\begin{equation}
h_\ell'
=
h_\ell
+
\gamma_\ell g_\ell
B\,
\operatorname{Read}
\left(
M_s^{(\ell)},q_\ell
\right).
\label{eq:inject}
\end{equation}
Here, $B$ maps the response back into the model's residual space. The gate
$g_\ell$ decides how strongly the KV-Skill should affect the current state, and
$\gamma_\ell$ controls the update magnitude.

The same projections $A$ and $B$ are shared across injection depths. Only the
gate and gain vary by layer. This restriction encourages the interface to read
the loaded KV-Skill rather than store an entire task by itself.

\begin{figure}[t]
    \centering
    \includegraphics[
        width=0.92\columnwidth,
        trim=8 6 8 6,
        clip
    ]{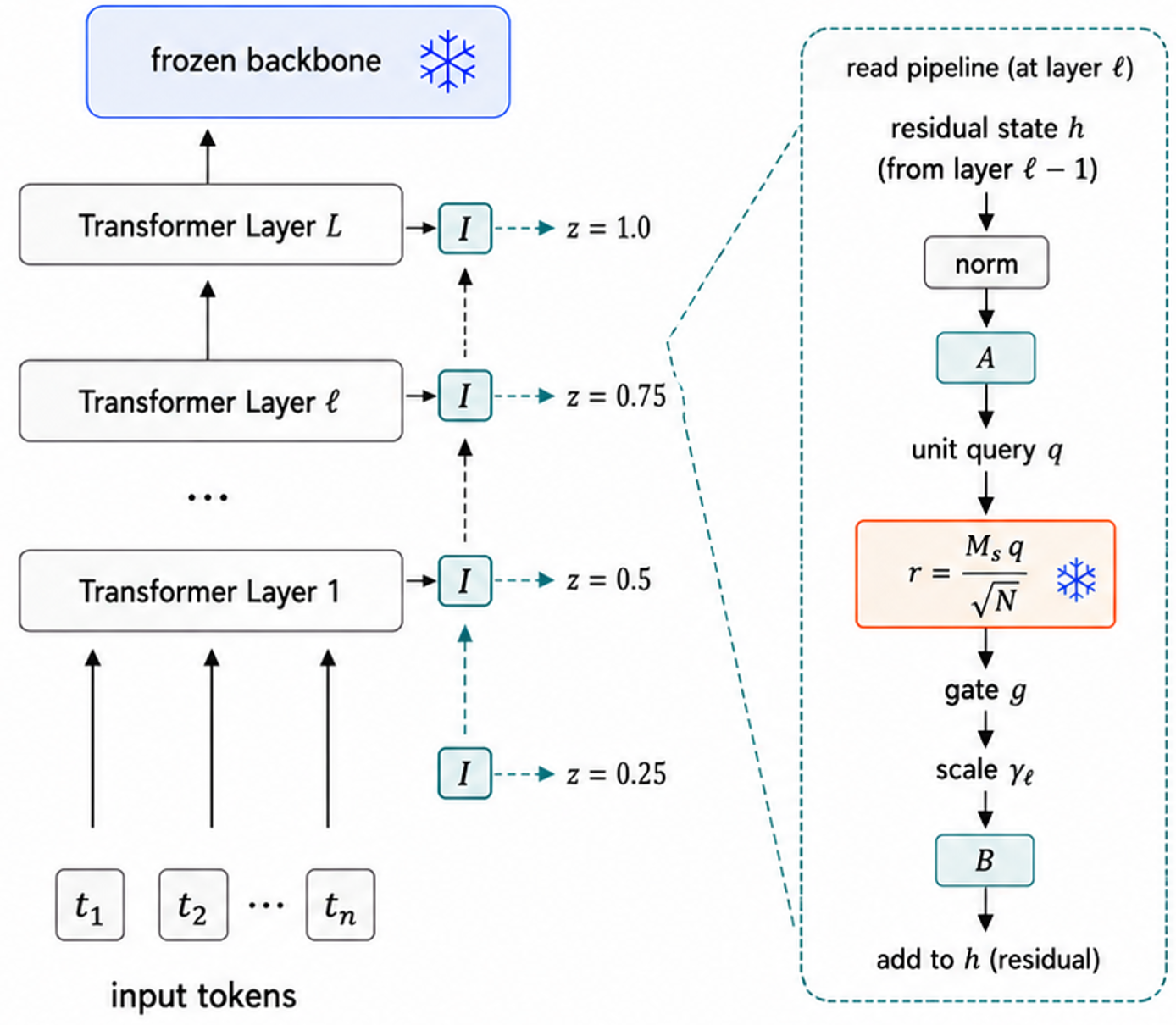}
    \caption{\textbf{Reading a KV-Skill.}
    The interface queries the loaded operator from the residual state and injects a gated response into the frozen backbone, separate from self-attention and without adding attention KV-cache entries.}
    \label{fig:KVskill_def}
\end{figure}

\subsection{Converting Text into a KV-Skill}
\label{sec:method-derive}

An authored text skill may contain useful knowledge even when the model cannot
reliably execute it from the prompt. We convert the document into a
text-derived KV-Skill using one frozen prefill.

At every injection layer, we record the residual states produced by the text
skill. Let
\begin{equation}
\bar H_s^{(\ell)}
\in
\mathbb{R}^{d_b\times N_s}
\end{equation}
contain the normalized states at layer $\ell$, where $N_s$ is the number of
text-skill tokens. We use the same states as both factors:
\begin{equation}
U_s^{(\ell)}
=
W_s^{(\ell)}
=
\bar H_s^{(\ell)}.
\label{eq:text-factors}
\end{equation}
The resulting operator is
\begin{equation}
M_s^{(\ell)}
=
\bar H_s^{(\ell)}
\bar H_s^{(\ell)\top}.
\label{eq:text-operator}
\end{equation}

This construction is self-associative. It measures how the current query aligns
with the text-skill states and returns a signed combination of those states.
The operator is derived separately at each injection depth because the
backbone represents the same document differently across layers.

Text-derived factors live in the residual space of the backbone that produced
them. We therefore derive a new text-derived $M_s$ for each backbone. The
source text and conversion procedure are shared, but the resulting tensors are
backbone-specific.

\subsection{Registering a Shared Interface}
\label{sec:method-reg}

Derivation creates the KV-Skill, but the backbone does not automatically know
how to use it. Registration teaches the interface $I$ to access the fixed
operator.

For each task input, we compare two conditions. The teacher reads the original
text skill in the prompt. The student removes the text skill and loads its
derived $M_s$. Registration combines teacher matching with gold-answer
supervision:
\begin{equation}
\mathcal{L}_{\mathrm{reg}}
=
\operatorname{KL}
\left(
p_{\mathrm{text}}
\,\Vert\,
p_{\mathrm{KV}}
\right)
+
\frac{1}{2}
\operatorname{CE}
\left(
p_{\mathrm{KV}},y
\right)
+
\lambda\Omega,
\label{eq:reg}
\end{equation}
where $\Omega$ controls the magnitude of the injected residual update.

Registration updates only the interface. The text-derived $M_s$ remains
unchanged. For each backbone, we register several KV-Skills sequentially into
one shared interface $I_b^{\mathrm{reg}}$. Previously registered KV-Skills are
replayed during this process to reduce forgetting. The resulting interface can
read different task-specific operators without storing a separate reader for
every text-derived KV-Skill.

\subsection{Learning a Compact KV-Skill from Reward}
\label{sec:method-rl}

Text conversion assumes that someone has already written a useful procedure.
Reward learning removes this requirement. It learns a compact KV-Skill directly
from task outcomes while keeping the backbone frozen.

The reward-learned construction uses one compact factor pair:
\begin{equation}
U_s,W_s
\in
\mathbb{R}^{d_s\times m_s},
\qquad
M_s=W_sU_s^\top.
\label{eq:compact-operator}
\end{equation}
The same operator is read at every injection depth. Unlike the text-derived
construction, $U_s$ and $W_s$ are independent and learned separately. In our
experiments, the operator contains 32 latent slots in a 256-dimensional skill
space.

Reward learning jointly optimizes the compact factors and a newly initialized
task interface:
\begin{equation}
\theta_{b,s}^{\mathrm{RL}}
=
\left\{
U_s,W_s,I_{b,s}^{\mathrm{RL}}
\right\}.
\label{eq:reward-parameters}
\end{equation}
The backbone remains frozen, and the prompt does not contain the text-skill
document.

We use verifier outcomes such as exact match or task success. For a group of
responses to the same input, we normalize their rewards:
\begin{equation}
\widehat{A}_i
=
\frac{
R_i-\operatorname{mean}(R_{1:G})
}{
\operatorname{std}(R_{1:G})+\epsilon
}.
\label{eq:adv}
\end{equation}
The normalized reward increases the probability of successful responses and
decreases the probability of unsuccessful ones. Complete optimization details
are provided in Appendix~\ref{sec:reward-details}.

We consider two initializations. \textsc{RL from scratch} begins without an
authored text skill. \textsc{RL from skill} begins from a compact
text-informed initialization. This initialization does not directly reuse the
backbone-specific, token-aligned factors in
Equation~\ref{eq:text-factors}; it initializes the compact reward-learning
construction.

\begin{table*}[t]
\centering
\footnotesize
\setlength{\tabcolsep}{3.6pt}
\renewcommand{\arraystretch}{1.0}
\begin{tabular}{@{}llccccc@{}}
\toprule
& & \multicolumn{3}{c}{\textbf{Main Tasks}}
  & \multicolumn{2}{c}{\textbf{General QA}} \\
\cmidrule(lr){3-5}\cmidrule(l){6-7}
\textbf{Backbone}
& \textbf{Method}
& \shortstack{\textbf{LiveMath}\\{\scriptsize Acc.}}
& \shortstack{\textbf{SearchQA}\\{\scriptsize EM}}
& \shortstack{\textbf{DocVQA}\\{\scriptsize ANLS}}
& \shortstack{\textbf{CSQA}\\{\scriptsize Acc.}}
& \shortstack{\textbf{OBQA}\\{\scriptsize Acc.}} \\
\midrule

\multirow[t]{8}{*}{\textit{Qwen3.5-4B}}
& Base                       & 32.3 & 68.1 & 86.9 & 80.7 & 90.6 \\
& Text Skill                 & 33.1 & 67.4 & 88.4 & 78.5 & 89.0 \\
& SkillOpt                   & 52.0 & 71.2 & 89.0 & --   & --   \\
& SoftSkill                  & 64.5 & 76.4 & 88.2 & --   & --   \\
\cmidrule(lr){2-7}
& KV-Skill (registered)      & 77.2\,{\scriptsize$\pm$2.5} & 79.3\,{\scriptsize$\pm$0.7} & -- & 79.3\,{\scriptsize$\pm$1.2} & 88.7\,{\scriptsize$\pm$0.8} \\
& KV-Skill (RL from skill)
  & \textbf{79.0}\,{\scriptsize$\pm$3.9}
  & \textbf{80.1}\,{\scriptsize$\pm$3.3}
  & \textbf{95.0}\,{\scriptsize$\pm$2.6}
  & -- & -- \\
& KV-Skill (RL from scratch)
  & 75.6\,{\scriptsize$\pm$4.1}
  & 80.0\,{\scriptsize$\pm$3.4}
  & 94.2\,{\scriptsize$\pm$2.9}
  & \textbf{81.1}\,{\scriptsize$\pm$1.6}
  & \textbf{90.8}\,{\scriptsize$\pm$2.2} \\
\midrule

\multirow[t]{6}{*}{\textit{Qwen3.6-35B-A3B}}
& Base                       & 33.9 & 72.7 & 87.6 & 86.3 & 96.2 \\
& Text Skill                 & 29.1 & 73.6 & 88.0 & 86.2 & 96.6 \\
& SkillOpt                   & 41.6 & 80.3 & 91.4 & --   & --   \\
& SoftSkill                  & 50.0 & 82.3 & 93.3 & --   & --   \\
\cmidrule(lr){2-7}
& KV-Skill (RL from skill)
  & 70.2\,{\scriptsize$\pm$4.4}
  & \textbf{82.8}\,{\scriptsize$\pm$1.1}
  & 95.4\,{\scriptsize$\pm$2.5}
  & \textbf{87.3}\,{\scriptsize$\pm$1.1}
  & 96.8\,{\scriptsize$\pm$1.5} \\
& KV-Skill (RL from scratch)
  & \textbf{72.6}\,{\scriptsize$\pm$4.3}
  & \textbf{82.8}\,{\scriptsize$\pm$1.2}
  & \textbf{95.6}\,{\scriptsize$\pm$2.4}
  & 87.1\,{\scriptsize$\pm$1.1}
  & \textbf{97.0}\,{\scriptsize$\pm$1.5} \\
\bottomrule
\end{tabular}
\caption{Main-task and general-QA results. Bold marks the best per backbone and column. SkillOpt and SoftSkill are reported on the main tasks.  Registered KV-Skill uses the SkillOpt-optimized skill. Other rows use the seed skill, the same one as Text Skill. (Appendix~\ref{sec:app-skills}).}
\label{tab:main}
\end{table*}

\section{Experiments}
\label{sec:exp}

Our experiments show that task knowledge need not remain in the form in which
it was acquired. A text procedure can be converted into a more effective
external operator, while an unwritten procedure can be discovered directly
from task reward. Under matched reward training, KV-Skill substantially
outperforms prefix-based representations on tasks requiring significant
adaptation and is competitive with LoRA and RL-trained steering vectors.
Surprisingly, a post-hoc SVD compresses a registered KV-Skill to one
task-aligned direction per injection layer while preserving nearly all its
benefit. A single interface can also serve eleven diverse KV-Skills at once with no
loss in average accuracy.

\label{sec:exp-overview}
We evaluate four backbones on ten benchmarks. The main-text tables cover
LiveMath \citep{he2026livemathematicianbench}, SearchQA
\citep{dunn2017searchqa}, DocVQA \citep{mathew2021docvqa}, CommonsenseQA
\citep{talmor2019commonsenseqa}, and OpenBookQA \citep{mihaylov2018can}.
STaRK-Prime and STaRK-MAG \citep{wu2024stark} and three agentic tasks are in
Appendix~\ref{sec:additional-results}, and Appendix~\ref{sec:exp-details} gives the full setup.

\subsection{Converting Text Skills to KV-Skills Improves Performance}
\label{sec:exp-registration}

Having the right procedure is not enough. The model must be able to use it. We derive $M_s$ from a text skill, keep it fixed, and train only
the shared interface $I$. This compares the same procedural knowledge as text
and as a KV-Skill.

On Qwen3.5-4B LiveMath, the registered KV-Skill comes
from the SkillOpt-optimized skill. Its matched text condition is therefore the
SkillOpt row of Table~\ref{tab:main}. The same document reaches 52.0 as text
and 77.2$\pm$2.5 as a KV-Skill. SoftSkill reaches 64.5, and the unoptimized seed skill 33.1.  Both
baselines learn from the same labeled training split. SkillOpt is a supervised
text baseline that edits the skill using verifier feedback on gold answers.
SoftSkill uses the same gold-answer cross-entropy as registration
(Appendix~\ref{sec:app-training}). With label access fixed, the bottleneck is
how the skill is represented. 

\begin{table}[t]
\centering
\scriptsize
\setlength{\tabcolsep}{3pt}
\renewcommand{\arraystretch}{1.0}
\begin{tabular}{@{}lcccc@{}}
\toprule
& \multicolumn{2}{c}{\textbf{LiveMath Acc.}} & \multicolumn{2}{c}{\textbf{SearchQA EM}} \\
\cmidrule(lr){2-3}\cmidrule(l){4-5}
\textbf{Prompt content} & No KV & +KV & No KV & +KV \\
\midrule
Empty (query only) & 32.3 & \textbf{77.2}{\tiny$\pm$2.5} & 68.1           & \textbf{79.3}{\tiny$\pm$0.7} \\
SkillOpt-optimized text skill & 52.0           & 76.6{\tiny$\pm$3.8} & 71.2           & 77.3{\tiny$\pm$2.4} \\
Same-length random tokens & 13.7{\tiny$\pm$3.1}   & 73.7{\tiny$\pm$0.5} & 66.3{\tiny$\pm$2.7}   & 79.2{\tiny$\pm$1.0} \\
\bottomrule
\end{tabular}
\caption{\textbf{Prompt content $\times$ registered KV-Skill} on Qwen3.5-4B.
+KV loads the operator derived from the SkillOpt-optimized skill. The first
two No-KV entries repeat
Table~\ref{tab:main}. Mean $\pm$ s.d.\ over three seeds.}
\label{tab:prompt-kv}
\end{table}

\begin{table*}[t]
\centering
\footnotesize
\setlength{\tabcolsep}{5.0pt}
\renewcommand{\arraystretch}{1.0}
\begin{tabular}{@{}llcccc@{}}
\toprule
& & \multicolumn{2}{c}{\textbf{Main Tasks}}
  & \multicolumn{2}{c}{\textbf{General QA}} \\
\cmidrule(lr){3-4}\cmidrule(l){5-6}
\textbf{Backbone}
& \textbf{Method}
& \shortstack{\textbf{LiveMath}\\{\scriptsize Acc.}}
& \shortstack{\textbf{SearchQA}\\{\scriptsize EM}}
& \shortstack{\textbf{CSQA}\\{\scriptsize Acc.}}
& \shortstack{\textbf{OBQA}\\{\scriptsize Acc.}} \\
\midrule

\multirow[t]{6}{*}{\textit{GLM-4.7-Flash}}
& Base                       & 18.6 & 66.3 & 76.5 & 87.4 \\
& Text Skill                 & 22.6 & 66.3 & 76.7 & 87.2 \\
& SkillOpt     & 40.3 & 70.0 & --   & --   \\
& SoftSkill     & 30.9 & 79.1 & --   & --   \\
\cmidrule(lr){2-6}
& KV-Skill (RL from skill)
  & \textbf{76.6}\,{\scriptsize$\pm$4.0}
  & 81.2\,{\scriptsize$\pm$2.4}
  & \textbf{78.5}\,{\scriptsize$\pm$1.8}
  & 88.6\,{\scriptsize$\pm$2.4} \\
& KV-Skill (RL from scratch)
  & 71.0\,{\scriptsize$\pm$4.3}
  & \textbf{81.8}\,{\scriptsize$\pm$2.3}
  & 77.6\,{\scriptsize$\pm$1.8}
  & \textbf{90.0}\,{\scriptsize$\pm$2.3} \\
\midrule

\multirow[t]{7}{*}{\textit{gpt-oss-20b}}
& Base                       & 22.5 & 65.1 & 81.6 & 94.6 \\
& Text Skill                 & 22.6	 & 64.5 & \textbf{83.1} & 94.8 \\
& SkillOpt     & 24.2 & 74.9 & --   & --   \\
& SoftSkill     & 28.2 & 66.1 & --   & --   \\
\cmidrule(lr){2-6}
& KV-Skill (registered)      & \textbf{65.3}\,{\scriptsize$\pm$0.8}  & \textbf{78.3}\,{\scriptsize$\pm$1.2} & 71.8\,{\scriptsize$\pm$2.8} & 82.3\,{\scriptsize$\pm$0.8} \\
& KV-Skill (RL from skill)
  & 54.8\,{\scriptsize$\pm$4.8}
  & 75.5\,{\scriptsize$\pm$3.7}
  & 81.5\,{\scriptsize$\pm$1.6} & 94.6\,{\scriptsize$\pm$1.2} \\
& KV-Skill (RL from scratch)
  & 52.6\,{\scriptsize$\pm$4.8}
  & 74.2\,{\scriptsize$\pm$3.8}
  & 81.6\,{\scriptsize$\pm$1.6}
  & \textbf{94.8}\,{\scriptsize$\pm$1.2} \\
\bottomrule
\end{tabular}
\caption{Additional backbones (no vision pathway, so no DocVQA). SkillOpt and SoftSkill are reproduced on the main tasks under their own protocols (Appendix~\ref{sec:app-training}). Registered KV-Skill uses the SkillOpt-optimized skill. Other rows use the seed skill, the same one as Text Skill.}
\label{tab:backbones}
\end{table*}

\paragraph{KV-Skill is a better representation of the skill than text.}
Table~\ref{tab:prompt-kv} varies the prompt while the registered operator is
absent or loaded. Adding the SkillOpt text back on top of its KV-Skill gives 76.6, so the text
adds nothing. The KV-Skill carries everything the text provides and makes it
usable. With random tokens of the same length in the prompt, the KV-Skill
still reaches 73.7, so the gain is not a prompt-length effect.
An identically supervised interface without the operator reaches only 71.7 on
SearchQA, against 77.3 with it (Appendix~\ref{sec:app-own-interface}). Registration also lifts STaRK-Prime from 41.6 to 49.7 and STaRK-MAG from 40.4
to 47.7 on Qwen3.5-4B (Appendix~\ref{sec:exp-stark}).

\subsection{Registration Distills a Task-Aligned Direction}
\label{sec:exp-rank}

The harvested operator starts with one slot per text token at every injection
depth, yet the registered behavior depends on a much simpler object. We
truncate each operator by SVD after registration, with the interface fixed. This post-hoc step yields an operator of about 20 KiB per skill (Appendix~\ref{sec:efficiency}). An analytic correction, $c_k=(k/N_s)^{1/4}$, keeps the read scale comparable
across ranks. One direction per injection layer matches the full
operator on LiveMath and STaRK-Prime and nearly matches it on SearchQA
(Figure~\ref{fig:rank-design}). With one registration seed, rank one and full rank both reach 79.8 on LiveMath
(Table~\ref{tab:app-rank}). The three-seed mean in Table~\ref{tab:main} is
77.2. An uncorrected sham control collapses, so this is not a scale artifact.

What survives is close to the mean residual direction of the text skill, and
it is task-aligned. Measured from Base, it keeps 90--100\% of the full operator's gain. A
magnitude-matched random direction fails on LiveMath (26.6) and STaRK-Prime
(43.0, against 41.6 for Base). The direction also
depends on the source text. At rank one, the SkillOpt-optimized skill reaches
79.8 and the seed skill 66.9 (Table~\ref{tab:app-source-doc}).

One direction per layer is enough, but applying it uniformly is not.
Query-dependent KV-Skill beats fixed steering by 25.0 points on LiveMath
(Table~\ref{tab:steering-control}). The compressed operator is therefore
\emph{text-derived, query-dependent conditional steering}, not a fixed
steering vector. This is one compressed point in the design space, chosen for efficiency. 
\begin{table}[t]
\centering
\small
\setlength{\tabcolsep}{7pt}
\renewcommand{\arraystretch}{1.0}
\begin{tabular}{@{}lc@{}}
\toprule
\textbf{Rank-one condition}
& \shortstack{\textbf{LiveMath}\\{\scriptsize Acc.}} \\
\midrule
Random direction                 & 26.6 \\
Fixed steering (learned $c$)     & 54.8 \\
Shuffled addressing              & 49.2  \\
Query-dependent KV-Skill         & \textbf{79.8} \\
\bottomrule
\end{tabular}
\caption{\textbf{Query-dependent addressing at rank one.}
Fixed steering uses a learned constant per layer instead of the query-dependent
coefficient. The random control matches update magnitude. One registration seed; Table~\ref{tab:main} reports the three-seed mean.}
\label{tab:steering-control}
\end{table}

\begin{table*}[t]
\centering
\footnotesize
\setlength{\tabcolsep}{5pt}
\renewcommand{\arraystretch}{1.0}
\newcommand{\std}[1]{\,{\scriptsize$\pm$#1}}
\begin{tabular}{@{}llrcccc@{}}
\toprule
& & & \multicolumn{2}{c}{\textbf{SearchQA EM}}
    & \multicolumn{2}{c}{\textbf{LiveMath Acc.}} \\
\cmidrule(lr){4-5}\cmidrule(l){6-7}
\textbf{Backbone} & \textbf{Trainable substrate} & \textbf{Params.}
& \textbf{Scratch} & \textbf{Skill init.}
& \textbf{Scratch} & \textbf{Skill init.} \\
\midrule
\multirow[t]{8}{*}{\textit{Qwen3.5-4B}}
& KV-Skill operator (RL)              & 1.330M
  & \textbf{80.0}\std{3.4} & \textbf{80.1}\std{3.3}
  & 75.6\std{4.1}          & \textbf{79.0}\std{3.9} \\
& Soft prefix (RL), $n{=}519$          & 1.329M
  & 74.0\std{2.3} & 73.8\std{2.9} & 59.7\std{3.1} & 68.6\std{3.1} \\
& Soft prefix (RL), $n{=}32$           & 81.9K
  & 77.3\std{0.8} & 71.8\std{5.6} & 59.7\std{2.8} & 64.0\std{2.8} \\
& Prefix KV (RL)                       & 1.327M
  & 69.7\std{0.3} & 71.7\std{0.9} & 16.1\std{2.1} & 28.0\std{3.8} \\
& LoRA (RL), $r{=}7$                   & 1.376M
  & 75.3\std{4.5} & --            & \textbf{78.6}\std{2.9} & -- \\
& LoRA (RL), $r{=}16$                  & 3.1M
  & 76.0\std{0.5} & --            & 66.9\std{4.1} & -- \\
& Bias-only steering (RL), 4 depths    & 10.2K
  & 78.9\std{0.9} & --            & 69.3\std{4.1} & -- \\
& Bias-only steering (RL), all 32 layers & 81.9K
  & 77.2\std{0.2} & --            & 66.9\std{5.7} & -- \\
\midrule
\multirow[t]{4}{*}{\textit{Qwen3.6-35B-A3B}}
& KV-Skill operator (RL)              & 1.067M
  & \textbf{82.8}\std{1.2} & \textbf{82.8}\std{1.1}
  & \textbf{72.6}\std{4.3} & \textbf{70.2}\std{4.4} \\
& Soft prefix (RL), $n{=}521$          & 1.067M
  & 73.2\std{1.8} & 73.3\std{1.3} & 28.8\std{3.3} & 26.3\std{8.5} \\
& Prefix KV (RL), $p{=}104$            & 1.065M
  & 74.4\std{0.2} & 75.2\std{0.4} & 33.1\std{4.3} & 25.5\std{2.8} \\
& LoRA (RL), $r{=}6$                   & 1.024M
  & 80.8\std{1.2} & --            & 52.7\std{10.9} & -- \\
\bottomrule
\end{tabular}
\caption{Matched reward training across trainable substrates. All methods
share the reward objective, backbone, training budget, and decoding protocol.
Parameter counts are matched, except for the smaller soft prefix ($n{=}32$),
the larger LoRA ($r{=}16$), and bias-only steering. Values are mean $\pm$
s.d.\ over three training seeds. Bold marks the best result per
backbone, task, and initialization. Skill init.\ uses the seed skill, the same one as Text Skill.}
\label{tab:substrate}
\end{table*}

\begin{table}[t]
\centering
\small
\setlength{\tabcolsep}{4.5pt}
\renewcommand{\arraystretch}{1.0}
\begin{tabular}{@{}llcc@{}}
\toprule
\textbf{Backbone}
& \textbf{Loaded KV-Skill}
& \shortstack{\textbf{LiveMath}\\{\scriptsize Acc.}}
& \shortstack{\textbf{SearchQA}\\{\scriptsize EM}} \\
\midrule
\multirow[t]{3}{*}{\textit{Qwen3.5-4B}}
& Random   & 25.4\,{\scriptsize$\pm$2.1}
               & 61.3\,{\scriptsize$\pm$0.9} \\
& Wrong task   & 36.3\,{\scriptsize$\pm$5.6}
               & 73.1\,{\scriptsize$\pm$2.4} \\
& Correct task & \textbf{77.2}\,{\scriptsize$\pm$2.5}
               & \textbf{79.3}\,{\scriptsize$\pm$0.7} \\
\midrule
\multirow[t]{3}{*}{\textit{gpt-oss-20b}}
& Random   & 24.7\,{\scriptsize$\pm$3.8}
               & 53.0\,{\scriptsize$\pm$3.8} \\
& Wrong task   & 27.2\,{\scriptsize$\pm$2.6} & 66.7\,{\scriptsize$\pm$5.7} \\
& Correct task & \textbf{65.3}\,{\scriptsize$\pm$0.8} & \textbf{78.3}\,{\scriptsize$\pm$1.2} \\
\bottomrule
\end{tabular}
\caption{Behavior follows the loaded KV-Skill when $I$ is fixed (mean
{\scriptsize$\pm$} s.d.\ over three registration seeds). Base is in
Tables~\ref{tab:main} and~\ref{tab:backbones}.}
\label{tab:identity}
\end{table}

\subsection{Task Behavior Resides in the Loaded KV-Skill}
\label{sec:exp-identity}

To test whether registration hides each task inside $I$, we hold $I$ fixed and
swap only the loaded $M_s$ for a random or wrong-task operator.

Behavior changes with the loaded operator (Table~\ref{tab:identity}).
Correct-task KV-Skills clearly beat wrong-task and random operators on both backbones. A separately fitted interface per operator gives the same result
(Appendix~\ref{sec:app-own-interface}). The interface provides access and the loaded $M_s$ decides the capability, so a
KV-Skill is loadable rather than hidden fine-tuning.

\subsection{Reward Optimization Improves and Discovers KV-Skills}
\label{sec:exp-reward-results}

Reward learning builds a KV-Skill directly from task feedback. We train a
compact operator and its task interface with the backbone frozen.
\textsc{RL from skill} starts from a text skill, and \textsc{RL from scratch}
starts from nothing. Both work across tasks and backbones
(Tables~\ref{tab:main} and~\ref{tab:backbones}).

\paragraph{Matched reward reveals a substrate effect.}
Table~\ref{tab:substrate} gives KV-Skill, soft prefix, prefix KV,
LoRA, and bias-only steering the same reward and training budget. KV-Skill is best or tied-best in seven of eight settings and leads every
comparison on the larger Qwen model. On Qwen3.5-4B, LoRA leads on LiveMath
(78.6$\pm$2.9 versus 75.6$\pm$4.1), and bias-only steering nearly ties on
SearchQA (78.9$\pm$0.9 versus 80.0$\pm$3.4). Both gaps are within seed
variance. Bias-only steering trails KV-Skill by 6--9 points on LiveMath
(Appendix~\ref{sec:app-steering}). Reward alone is not enough. The substrate matters too.

\paragraph{More reward-learning results.}
Behavior cloning and capacity controls are in
Appendix~\ref{sec:app-reward-controls}. On agentic tasks, reward learning beats
SkillOpt on OfficeQA and SpreadsheetBench but not on ALFWorld
(Appendix~\ref{sec:app-agentic}).

\subsection{One Interface Serves Many Skills}
\label{sec:exp-retention}

The registered Qwen3.5-4B results above already come from one interface shared
by nine skills (Appendix~\ref{sec:app-registration}).
Table~\ref{tab:app-retention-ext} goes further: we register eleven diverse
skills one at a time with replay, spanning math,
retrieval, science, multi-hop, and numerical reading. First, there is no loss on
average. After all eleven, the shared interface averages 78.1, against 77.8
when each skill has its own interface, and matches or beats it on seven
skills. Second, registered skills stay stable as new ones arrive. SearchQA
stays between 78.7 and 81.6 over ten later stages, and SciQ between 94.7 and
95.3. The one clear cost is LiveMath, which ends at 73.4 against 76.9 on its
own interface. Third, sharing can help. STaRK-Prime gains 6.8 points, possibly
because more skills reduce overfitting (Appendix~\ref{sec:app-retention}). The
interface also does not grow with the library (Table~\ref{tab:interface-cost}).
Adding a skill costs only its operator and one registration run, so one
interface can serve a growing skill library.

\subsection{Discussion: The LiveMath Results}
\label{sec:exp-discussion}

\paragraph{The baselines are genuine.}
We evaluate Base and Text Skill with a budget large enough that no response in Tables~\ref{tab:main} and~\ref{tab:backbones} is cut off or truncated
(Appendix~\ref{sec:app-protocol}). On Qwen3.5-4B, our Base matches the
no-skill scores that SkillOpt and SoftSkill report on SearchQA (68.1) and
DocVQA (86.9) \citep{yang2026skillopt,tao2026softskill}. On LiveMath, both
papers report 22.4, below our 32.3. The task is hard even for frontier models:
SkillOpt reports 37.6 for GPT-5.5 without a skill. So neither truncation nor answer length explains the gap to 77.2.

\paragraph{LiveMath is a threshold task.}
The optimized LiveMath skill encodes one procedure: rank the options by
theorem strength and check exact conditions (Appendix~\ref{sec:app-skills}).
Accuracy jumps once a model applies it. In
SkillOpt's evaluation on Qwen3.5-4B, human-written, one-shot,
trajectory-mined, and prompt-optimized skills all stay between 10.6 and 28.8.
SkillOpt's LiveMath gain on GPT-5.5 also comes from one accepted edit. Methods that train on the task find the procedure and jump.
SkillOpt, a supervised text baseline, reaches 52.0, and SoftSkill reaches
64.5. Our registered KV-Skill is built from SkillOpt's optimized skill, so it
inherits the same procedure. Registration makes the procedure usable and
reaches 77.2. Reward learning finds it without any text and reaches 75.6. The gain also does not come from the interface
just learning the answer format. With the same interface, a random or
wrong-task operator scores only 25.4 or 36.3 (Table~\ref{tab:identity}).

\section{Conclusion}
\label{sec:conclusion}

KV-Skill keeps task knowledge outside the prompt and backbone, in a design
space with two paths. Registration represents a text skill better than text, and reward learning
builds one from experience.

\section{Limitations}
\label{sec:limitations}

KV-Skill moves task knowledge outside the prompt and backbone, but it does not
eliminate deployment costs. Reading the operator adds computation at each
injection depth, and every backbone requires an interface. Text-derived
KV-Skills are also backbone-specific and can require substantial storage before
compression. Loading a KV-Skill after an interaction has begun may require
prefilling the existing history again if earlier states must be reinterpreted.

Our experiments cover several tasks and backbone families, but the most detailed
analyses focus on Qwen3.5-4B and QA-style benchmarks. Evidence on long-horizon
agentic tasks remains limited, where sparse episode-level rewards make credit
assignment difficult. We therefore do not claim that the observed gains or
rank-one compression will generalize to all tasks, models, or interaction
settings.

Finally, we do not evaluate direct transfer of the same KV-Skill tensor across
backbones. Text-derived KV-Skills are constructed in each backbone's residual
space, while the portability of compact reward-learned KV-Skills remains an
open question.

\section{Ethical Considerations}
KV-Skill is a general mechanism for developing and loading task behavior.
A KV-Skill could therefore encode harmful or misleading procedures, and reward
learning could exploit weaknesses in an imperfect verifier. External
KV-Skills may also make behavioral changes less visible than instructions
written in the prompt. Practical deployments should verify skill provenance,
evaluate each KV-Skill before use, restrict access to untrusted operators, and
retain mechanisms for logging, removal, and rollback.

KV-Skill also inherits the biases and safety limitations of its underlying
backbone. Although the backbone remains frozen, registration and reward
optimization still require GPU computation, particularly for rollout
generation, and therefore have a nonzero environmental cost.
\paragraph{Use of AI assistants.}
AI assistants were used to support manuscript organization and language
revision, as well as code implementation and debugging. The authors reviewed
and modified all generated suggestions, verified the experimental results and
citations, and take full responsibility for the final manuscript and released
artifacts.








\newpage
\bibliography{custom}

\appendix
\raggedbottom

\section{KV-Skill Construction and Training}
\label{sec:app-method}

\subsection{Two KV-Skill Constructions}
\label{sec:app-skill-design}

Table~\ref{tab:skill-design} summarizes the two constructions.

\begin{table}[H]
\centering
\footnotesize
\setlength{\tabcolsep}{3pt}
\renewcommand{\arraystretch}{1.08}
\begin{tabular}{@{}p{0.25\columnwidth}p{0.34\columnwidth}p{0.34\columnwidth}@{}}
\toprule
\textbf{Property}
& \textbf{Text-derived KV-Skill}
& \textbf{Reward-learned KV-Skill} \\
\midrule
Knowledge source
& Authored text skill
& Verifier reward \\

Representation
& Token-aligned states in the backbone's residual space
& Compact latent skill factors \\

Layer organization
& Separate operator at each injection depth
& One operator shared across depths \\

Optimization
& $M_s$ remains fixed; registration updates the shared interface
& $M_s$ and a task interface are jointly optimized \\

Primary role
& Compile existing task knowledge
& Discover task behavior from experience \\
\bottomrule
\end{tabular}
\caption{Two constructions within the KV-Skill design space. Both use the same
associative read but differ in how the operator is obtained and optimized.}
\label{tab:skill-design}
\end{table}

\subsection{Registration Protocol}
\label{sec:app-registration}

Each skill is registered for 300 AdamW steps (learning rate $2{\times}10^{-4}$)
with the loss in Equation~\ref{eq:reg} ($\lambda=0.3$), replaying previously
registered skills. The nine-skill and eleven-skill interfaces below start from
the same interface, pre-trained on universal skills before any benchmark skill
is registered. Each sequence is run with three registration seeds.

\paragraph{Nine-skill interface.}
The registered Qwen3.5-4B results in Tables~\ref{tab:main},
\ref{tab:steering-control}, and~\ref{tab:identity} and the LiveMath and
SearchQA rank analysis use one interface registered on nine skills in this
order: SearchQA, LiveMath, CommonsenseQA, OpenBookQA, ARC
\citep{clark2018thinksolvedquestionanswering}, SciQ \citep{welbl-etal-2017-crowdsourcing}, QASC
\citep{Khot_2020}, BoolQ \citep{clark-etal-2019-boolq}, and GPQA
\citep{rein2024gpqa}. ARC, SciQ, QASC, BoolQ, and GPQA are registered to
populate the shared interface and are not used as evaluation benchmarks.
Table~\ref{tab:interface-cost} reports the size of this interface. The
registered Qwen3.5-4B STaRK results in Table~\ref{tab:stark} add the STaRK
skill to this interface after the nine skills.

\paragraph{Retention sequences.}
Two further sequences study retention and are separate from the nine-skill
interface. The three-skill sequence in Section~\ref{sec:exp-retention}
registers SearchQA, LiveMath, and STaRK-Prime. The eleven-skill sequence
registers SearchQA, LiveMath, STaRK-Prime, CommonsenseQA, OpenBookQA, ARC,
SciQ, QASC, BoolQ, HotpotQA
\citep{yang-etal-2018-hotpotqa}, and DROP \citep{dua-etal-2019-drop};
Table~\ref{tab:app-retention-ext} reports all of them.

\subsection{Reward Optimization Protocol}
\label{sec:reward-details}

Reward learning freezes the backbone and updates the compact factors together
with a newly initialized task interface. For each input, we sample a group of
responses, score them with the task verifier, and normalize rewards within the
group as in Equation~\ref{eq:adv}. Groups with constant rewards are skipped.
\textsc{RL from scratch} empty initializes the compact factors.
\textsc{RL from skill} uses a compact text-informed initialization before the
same reward optimization. The matched-substrate experiments use the same training and decoding protocol for KV-Skill, soft prefix, prefix
KV, LoRA, and bias-only steering;
only the trainable substrate and its structurally available initialization
change.

Unless noted, reward optimization runs for 300 steps with 8 prompts per step
and a group size of 8 responses per prompt, sampling at temperature 1.0. We
optimize with AdamW (weight decay 0), clip each tensor to Frobenius norm 1.0,
cap the truncated importance-sampling ratio at 2.0, and process one sequence
per learner micro-batch.

\section{Experimental Setup}
\label{sec:exp-details}

\subsection{Tasks, Splits, and Metrics}

Our main evaluation contains three tasks. LiveMath tests mathematical reasoning
and is scored by accuracy. SearchQA tests retrieval-based question answering
and is scored by exact match (EM). DocVQA tests document question answering and
is scored by average normalized Levenshtein similarity (ANLS). We use
CommonsenseQA (CSQA) and OpenBookQA (OBQA) to measure general
question-answering ability.

We additionally study structured retrieval and agentic behavior. STaRK-Prime
and STaRK-MAG evaluate retrieval over semi-structured knowledge graphs. We
reduce each task to candidate selection and score it by exact match
(Appendix~\ref{sec:exp-stark}). OfficeQA \citep{opsahl2026officeqa},
SpreadsheetBench \citep{ma2024spreadsheetbench}, and ALFWorld
\citep{shridhar2021alfworld} evaluate long-horizon agentic behavior and are
scored by task success rate.

Each benchmark is scored with its native evaluator. Dataset-backed benchmarks
use deterministic train/selection/test splits with a 2:1:7 ratio and split seed
42; the selection split is used only for checkpoint selection, and all reported
scores are computed on the held-out test split. The only exception is the
single-skill LiveMath baseline, whose selection split coincides with its test
split. The test splits contain 1,400 SearchQA, 124 LiveMath, 374 DocVQA, 1,221
CSQA, 500 OBQA, 500 STaRK-Prime, 497 STaRK-MAG, 172 OfficeQA, 280
SpreadsheetBench, and 134 ALFWorld examples. Registered KV-Skill results are
averaged over three registration seeds on seeded test subsets (SearchQA 300,
CSQA and OBQA 200, and all 124 LiveMath examples).

\subsection{Backbones, Decoding, and Answer Extraction}
\label{sec:app-protocol}

We evaluate Qwen3.5-4B, Qwen3.6-35B-A3B, GLM-4.7-Flash, and gpt-oss-20b.
Thinking mode is disabled through the chat template on Qwen and GLM
backbones. gpt-oss-20b cannot disable its reasoning channel; we use its lowest
available reasoning effort for registered evaluations and the template default
for reward-learned evaluations. No Base or Text Skill response in Tables~\ref{tab:main}
and~\ref{tab:backbones} is cut off by the generation budget. Answers are parsed with the benchmarks' own scorers
\citep{yang2026skillopt}. A response is scored up to its first
\texttt{</answer>}; when no tag is present, the scorer falls back to the last
line.

\subsection{Compared Methods and Training Signals}
\label{sec:app-training}

\textsc{Base} uses the unmodified backbone, and \textsc{Text Skill} places the
original text skill in the prompt. \textsc{SkillOpt} \citep{yang2026skillopt}
edits the skill text with an external optimizer that reflects on
verifier-scored rollouts of the target model and accepts an edit only if it
passes a validation gate. \textsc{SoftSkill} \citep{tao2026softskill} distills
the skill into a 32-vector continuous prefix trained with next-token
cross-entropy on the gold answer \texttt{<answer>$y$</answer>}, which is the
same gold-answer term used in our registration loss.

\textsc{KV-Skill (registered)} converts a text skill into the KV-Skill $M_s$
with one prefill pass and registers $M_s$ through the model-specific interface
$I$. Registration updates $I$ but never changes $M_s$.
\textsc{KV-Skill (RL from skill)} starts reward optimization from a
KV-Skill converted from a text skill, whereas
\textsc{KV-Skill (RL from scratch)} starts from an empty KV-Skill
initialization (Section~\ref{sec:method-rl}).

Apart from \textsc{Text Skill}, every compared method learns from the labeled
training split of the same 2:1:7 split, either through a verifier or through
gold answers; the baselines therefore have supervised access to the task.

On GLM-4.7-Flash and gpt-oss-20b (Table~\ref{tab:backbones}), SkillOpt and
SoftSkill are reproduced with their released code under their own decoding
protocols, which differ from ours. SkillOpt uses temperature 0.7 and 8,000
tokens, with thinking off on GLM-4.7-Flash and reasoning on for gpt-oss-20b.
SoftSkill uses greedy decoding with 16 (LiveMath) or 64 (SearchQA) new tokens
and thinking off, with the final channel forced on gpt-oss-20b. SoftSkill is
the mean $\pm$ standard deviation over three seeds. SkillOpt is a single run
reported at its best selection step with a binomial standard error, following
its own convention.

\subsection{Skill Documents and Prompt Templates}
\label{sec:app-skills}

\begin{figure*}[t]
\centering
\scriptsize
\begin{verbatim}
You are an expert question answering agent.

{skill_section}## Task Format
You will receive a CONTEXT containing document passages and a QUESTION.
Read the context carefully and answer the question based on the information provided.

## Answer Format
Think step by step, then provide your final answer inside <answer>...</answer> tags.
Keep your answer concise - typically a few words or a short phrase.
Do not repeat the question. Do not include unnecessary explanation in the answer tags.

Example:
<answer>Abraham Lincoln</answer>
\end{verbatim}
\caption{SearchQA system prompt template. The \texttt{\{skill\_section\}}
placeholder receives the skill document in the \textsc{Text Skill} condition and
is empty in all KV-Skill conditions.}
\label{fig:app-prompt}
\end{figure*}

Registered KV-Skills on SearchQA and LiveMath are converted from
SkillOpt's GPT-5.5-optimized skill \citep{yang2026skillopt}; on other tasks,
Text Skill and registration share one authored document. The SearchQA seed
skill contains no rules. We reproduce two documents below, followed by the task
prompt template into which a skill is inserted. The KV-Skill conditions use the
same template with an empty skill section.

\paragraph{LiveMath skill document:}\mbox{}\\
\textbf{Live Mathematical MCQ Heuristics}

\textit{Option Comparison}
\begin{itemize}\itemsep0pt \parsep0pt \topsep2pt \partopsep0pt
\item Compare all options before committing. The correct choice is often the
strongest statement justified by the question, while nearby distractors are
weaker, overstrong, or miss an equality case.
\item Track exact quantifiers such as ``there exists'', ``for every'',
``if and only if'', and ``exactly when''.
\end{itemize}

\textit{Theorem-Level Precision}
\begin{itemize}\itemsep0pt \parsep0pt \topsep2pt \partopsep0pt
\item Check whether an option weakens the conclusion by dropping a
characterization, equality clause, or full equivalence.
\item Check whether an option overstates the theorem by upgrading regularity,
removing scale restrictions, or changing an existential statement into a
universal one.
\end{itemize}

\textit{Hypotheses}
\begin{itemize}\itemsep0pt \parsep0pt \topsep2pt \partopsep0pt
\item Verify the hypotheses and domain carefully. Distractors often keep the
theorem shape but alter the required assumptions.
\item Pay close attention to equality cases, extremal conditions, and whether a
result applies to the full family or only a restricted subfamily.
\end{itemize}

\textit{Final Answer}
\begin{itemize}\itemsep0pt \parsep0pt \topsep2pt \partopsep0pt
\item Output the final answer as the single option label only.
\end{itemize}

\paragraph{DocVQA skill document:}\mbox{}\\
\textbf{DocVQA Skill}

\textit{Visual Evidence Discipline}
\begin{itemize}\itemsep0pt \parsep0pt \topsep2pt \partopsep0pt
\item Read the document carefully before answering.
\item Prefer the smallest exact text span that answers the question.
\item When several nearby strings look similar, choose the one whose
surrounding labels or layout best match the question.
\end{itemize}

\textit{Exact Answer Discipline}
\begin{itemize}\itemsep0pt \parsep0pt \topsep2pt \partopsep0pt
\item Copy names, numbers, and dates exactly from the document whenever
possible.
\item Prefer direct extraction over paraphrase.
\item Before finalizing, compare the answer against nearby alternatives and
keep the best-supported exact span.
\end{itemize}

\section{Additional Experimental Results}
\label{sec:additional-results}

The results below follow the order of Section~\ref{sec:exp}.
Appendix~\ref{sec:exp-stark} extends Section~\ref{sec:exp-registration},
Appendix~\ref{sec:app-rank} extends
Section~\ref{sec:exp-rank}, Appendix~\ref{sec:app-own-interface} extends
Section~\ref{sec:exp-identity}, Appendix~\ref{sec:app-reward-controls} extends
Section~\ref{sec:exp-reward-results}, and Appendix~\ref{sec:app-retention}
extends Section~\ref{sec:exp-retention}. Appendix~\ref{sec:app-agentic} reports
long-horizon agentic tasks.

\begin{figure*}[t]
    \centering
    \includegraphics[width=0.92\textwidth]{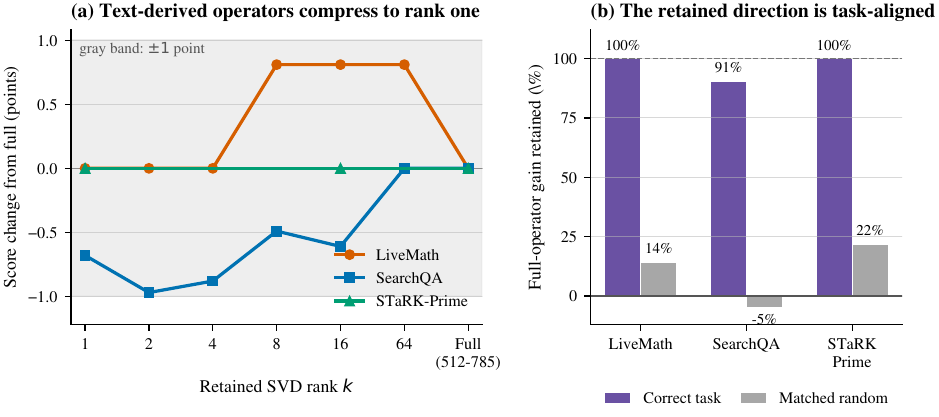}
    \caption{\textbf{Post-hoc rank and direction analysis.}
    Left: spectral truncation preserves full-operator performance down to rank
    one while the registered interface remains fixed. Right: the correct
    rank-one direction preserves nearly all of the gain, whereas a
    magnitude-matched random direction does not. Results use 124 LiveMath,
    300 SearchQA, and 100 STaRK-Prime examples.}
    \label{fig:rank-design}
\end{figure*}

\subsection{Knowledge-Graph Retrieval}
\label{sec:exp-stark}

Knowledge-graph retrieval exposes the same access problem in a different form.
An agent must apply the graph schema and retrieval procedure while reasoning
over query evidence. We evaluate this ability through ten-candidate selection
on STaRK-Prime and STaRK-MAG.

Text Skill remains near Base, while registration improves both graphs on both
backbones (Table~\ref{tab:stark}). RL from scratch also improves retrieval
without receiving an authored schema or procedure. Authored graph knowledge is
most valuable after conversion, but reward can recover useful behavior when
that knowledge is unavailable. Structured retrieval therefore brings the two
paths together: one unlocks existing knowledge, while the other discovers part
of the procedure from experience.

\begin{table}[H]
\centering
\small
\setlength{\tabcolsep}{9pt}
\renewcommand{\arraystretch}{1.10}
\begin{tabular}{@{}lcc@{}}
\toprule
\textbf{Method}
& \shortstack{\textbf{Prime}\\{\scriptsize EM}}
& \shortstack{\textbf{MAG}\\{\scriptsize EM}} \\
\midrule

\multicolumn{3}{@{}l}{\textit{Qwen3.5-4B}} \\
Base                       & 41.6 & 40.4 \\
Text Skill                 & 39.8 & 40.0 \\
\cmidrule(lr){1-3}
KV-Skill (registered)      & \textbf{49.7} & \textbf{47.7} \\
KV-Skill (RL from scratch)
  & 44.2\,{\scriptsize$\pm$1.1}
  & 43.1\,{\scriptsize$\pm$1.5} \\
\midrule

\multicolumn{3}{@{}l}{\textit{gpt-oss-20b}} \\
Base                       & 51.7 & 38.3 \\
Text Skill                 & 54.0 & 37.7 \\
\cmidrule(lr){1-3}
KV-Skill (registered)      & \textbf{56.3}  & \textbf{46.7} \\
KV-Skill (RL from scratch) & 53.3\,{\scriptsize$\pm$1.5}  & 44.85\,{\scriptsize$\pm$2.5} \\
\bottomrule
\end{tabular}
\caption{STaRK candidate-selection results on Qwen3.5-4B and gpt-oss-20b.
Each example contains ten candidates and is evaluated by exact match.
Bold marks the best result within each backbone and graph. On Qwen3.5-4B,
registration adds the STaRK skill after the nine skills of
Appendix~\ref{sec:app-registration}.}
\label{tab:stark}
\end{table}

\subsection{Rank and Direction Controls}
\label{sec:app-rank}

Table~\ref{tab:app-rank} gives the numerical values behind
Figure~\ref{fig:rank-design}. Spectral truncation is applied after registration
while the interface remains fixed. The rank-dependent correction
$c_k=(k/N_s)^{1/4}$ is derived analytically and is not fitted to task accuracy.

\begin{table}[H]
\centering
\small
\setlength{\tabcolsep}{5pt}
\renewcommand{\arraystretch}{1.08}
\begin{tabular}{@{}rccc@{}}
\toprule
\textbf{Rank}
& \shortstack{\textbf{LiveMath}\\{\scriptsize Acc.}}
& \shortstack{\textbf{SearchQA}\\{\scriptsize EM}}
& \shortstack{\textbf{STaRK-Prime}\\{\scriptsize EM}} \\
\midrule
1    & 79.84 & 78.33 & 48.00 \\
4    & 79.84 & 78.33 & -- \\
16   & 80.65 & 78.33 & 48.00 \\
64   & 80.65 & 79.33 & -- \\
Full & 79.84 & 79.33 & 48.00 \\
\bottomrule
\end{tabular}
\caption{Post-hoc rank truncation of registered text-derived KV-Skills (one registration seed; 124 LiveMath, 300 SearchQA, and 100 STaRK-Prime examples). Table~\ref{tab:main} reports the three-seed LiveMath mean (77.2). Full
rank is 512 for LiveMath and SearchQA and 785 for STaRK-Prime.}
\label{tab:app-rank}
\end{table}

The correction keeps the relative injected-update magnitude nearly constant,
from 0.0233 at rank one to 0.0255 at full rank. An uncorrected sham control
instead increases this ratio to 0.9808 at rank one and collapses LiveMath
accuracy to 0.8. At full rank, the corrected and sham constructions reduce to
the original operator and agree in both magnitude and accuracy. 

The retained rank-one direction also depends on the source text
(Table~\ref{tab:app-source-doc}). Under the same registered interface, the
rank-one operator derived from the SkillOpt-improved skill reaches 79.84 on
LiveMath, whereas the one derived from the seed skill reaches 66.9.

\begin{table}[H]
\centering
\small
\setlength{\tabcolsep}{8pt}
\renewcommand{\arraystretch}{1.08}
\begin{tabular}{@{}lc@{}}
\toprule
\textbf{LiveMath source} & \textbf{Rank-1 Acc.} \\
\midrule
Primary skill & 79.84 \\
Seed skill (same as Text skill)    & 66.9 \\
\bottomrule
\end{tabular}
\caption{Source-document control at rank one on Qwen3.5-4B LiveMath (one registration seed). Primary
skill is the SkillOpt-improved skill used for registration; seed skill is the
SkillOpt seed skill. Both operators are truncated to rank one and read through
the same registered interface.}
\label{tab:app-source-doc}
\end{table}

\subsection{Separately Fitted Interfaces}
\label{sec:app-own-interface}

Table~\ref{tab:identity} holds the registered interface fixed while swapping
the operator. To test whether the drop reflects an interface that cannot read
an unfamiliar operator, we instead fit a separate interface to each operator
with identical SearchQA supervision: the same starting interface, data, loss,
and 300 registration steps. Only the correct-task operator gives a clear gain with its own fitted
interface: 77.3, against 71.7 without an operator, 71.0 with the wrong-task
operator, and 73.0 with random tokens (Table~\ref{tab:app-own-interface}).

\begin{table}[H]
\centering
\small
\setlength{\tabcolsep}{8pt}
\renewcommand{\arraystretch}{1.08}
\begin{tabular}{@{}lc@{}}
\toprule
\textbf{Operator} & \textbf{SearchQA EM} \\
\midrule
None                      & 71.7 \\
Correct (SearchQA)        & 77.3 \\
Wrong-task (LiveMath)     & 71.0 \\
Random tokens             & 73.0 \\
\bottomrule
\end{tabular}
\caption{Interface control on Qwen3.5-4B SearchQA ($n{=}300$). Each operator
receives its own interface, registered from the same starting point with the
same data, loss, and number of steps; the registration teacher reads the
document from which that operator is derived. Random tokens match the
SearchQA skill in token count.}
\label{tab:app-own-interface}
\end{table}

\subsection{Reward-Learning Controls}
\label{sec:app-reward-controls}

\paragraph{Behavior-cloning controls.}
\label{sec:app-bc}

Table~\ref{tab:app-bc} reports behavior cloning with the same frozen
backbones. On the 4B model, behavior cloning stays within 1.6 points of Base on SearchQA
(68.1) and gives no gain on LiveMath. The larger model can benefit on some tasks, but the
pattern differs from reward optimization, especially on LiveMath.

\begin{table}[H]
\centering
\scriptsize
\setlength{\tabcolsep}{2.6pt}
\renewcommand{\arraystretch}{1.08}
\begin{tabular}{@{}lcccccc@{}}
\toprule
& \multicolumn{2}{c}{\textbf{SearchQA EM}}
& \multicolumn{2}{c}{\textbf{LiveMath Acc.}}
& \multicolumn{2}{c}{\textbf{DocVQA ANLS}} \\
\cmidrule(lr){2-3}\cmidrule(lr){4-5}\cmidrule(l){6-7}
\textbf{Substrate}
& \textbf{Scr.} & \textbf{Skill}
& \textbf{Scr.} & \textbf{Skill}
& \textbf{Scr.} & \textbf{Skill} \\
\midrule
\multicolumn{7}{@{}l}{\textit{Qwen3.5-4B}} \\
KV-Skill     & 68.36 & 68.93 & 20.16 & 21.77 & 93.82 & 94.18 \\
Soft prefix  & 67.21 & 67.00 & 24.19 & 19.35 & 92.88 & 93.66 \\
Prefix KV    & 69.64 & 66.93 & 17.74 & 18.55 & 92.35 & 93.78 \\
LoRA         & 69.14 & --    & 19.35 & --    & 93.68 & -- \\
\midrule
\multicolumn{7}{@{}l}{\textit{Qwen3.6-35B-A3B}} \\
KV-Skill     & 84.29 & 82.50 & 57.26 & 58.87 & 93.61 & 95.35 \\
Soft prefix  & 81.57 & 79.64 & 25.81 & 32.26 & 95.42 & 95.49 \\
Prefix KV    & 74.50 & 77.29 & 24.19 & 21.77 & 95.25 & 95.58 \\
LoRA         & 82.07 & --    & 62.90 & --    & 96.90 & -- \\
\bottomrule
\end{tabular}
\caption{Behavior-cloning controls. \textbf{Scr.} is scratch initialization and
\textbf{Skill} is skill initialization. A dash denotes an initialization that is
not structurally available.}
\label{tab:app-bc}
\end{table}

\paragraph{Compact KV-Skill capacity.}
\label{sec:app-capacity}

We vary the slot count $m_s$ while holding $d_s=256$, and vary $d_s$ while
holding $m_s=32$. These are single-run capacity sweeps on Qwen3.5-4B SearchQA;
the production setting is reported with its main experimental estimate.

\begin{table}[H]
\centering
\small
\setlength{\tabcolsep}{6pt}
\renewcommand{\arraystretch}{1.08}
\begin{tabular}{@{}rcc@{}}
\toprule
\textbf{Slots $m_s$} & \textbf{Scratch} & \textbf{Skill init.} \\
\midrule
1   & 77.86 & 78.21 \\
4   & 80.14 & 78.93 \\
32  & 80.00 & 80.07 \\
64  & 78.64 & 80.43 \\
128 & 77.43 & -- \\
\midrule
\textbf{Dimension $d_s$} & \textbf{Scratch} & \textbf{Skill init.} \\
\midrule
64  & 78.93 & 80.21 \\
128 & 78.79 & 79.14 \\
256 & 80.00 & 80.07 \\
512 & 80.86 & 80.50 \\
\bottomrule
\end{tabular}
\caption{SearchQA EM under compact KV-Skill capacity changes. The slot sweep
changes $m_s$ by 128-fold; the dimension sweep changes the dominant parameter
budget by eight-fold.}
\label{tab:app-capacity}
\end{table}

\begin{table*}[t]
\centering
\scriptsize
\setlength{\tabcolsep}{2.1pt}
\renewcommand{\arraystretch}{1.08}
\begin{tabular}{@{}lccccccccccc@{}}
\toprule
\textbf{Stage} & \textbf{SearchQA} & \textbf{LiveMath} & \textbf{STaRK-Prime}
& \textbf{CSQA} & \textbf{OBQA} & \textbf{ARC} & \textbf{SciQ} & \textbf{QASC}
& \textbf{BoolQ} & \textbf{HotpotQA} & \textbf{DROP} \\
\midrule
After SearchQA & 77.2$\pm$0.7 & -- & -- & -- & -- & -- & -- & -- & -- & -- & -- \\
After LiveMath & 79.7$\pm$1.7 & 65.1$\pm$2.0 & -- & -- & -- & -- & -- & -- & -- & -- & -- \\
After STaRK-Prime & 80.8$\pm$1.2 & 75.8$\pm$0.8 & 39.3$\pm$3.7 & -- & -- & -- & -- & -- & -- & -- & -- \\
After CSQA & 79.8$\pm$2.7 & 77.2$\pm$2.3 & 45.3$\pm$4.9 & 81.5$\pm$0.5 & -- & -- & -- & -- & -- & -- & -- \\
After OBQA & 81.6$\pm$1.7 & 75.3$\pm$2.8 & 42.8$\pm$2.9 & 81.7$\pm$0.6 & 88.5$\pm$0.9 & -- & -- & -- & -- & -- & -- \\
After ARC & 81.3$\pm$1.2 & 75.3$\pm$1.2 & 46.9$\pm$2.0 & 81.5$\pm$1.5 & 88.7$\pm$1.3 & 92.2$\pm$0.3 & -- & -- & -- & -- & -- \\
After SciQ & 80.1$\pm$2.0 & 74.2$\pm$4.2 & 46.3$\pm$1.7 & 79.5$\pm$1.0 & 88.8$\pm$0.3 & 91.8$\pm$0.3 & 95.0$\pm$0.5 & -- & -- & -- & -- \\
After QASC & 80.8$\pm$2.5 & 75.3$\pm$1.2 & 46.7$\pm$3.5 & 79.7$\pm$2.8 & 87.8$\pm$0.6 & 92.2$\pm$0.6 & 95.3$\pm$0.3 & 80.8$\pm$1.0 & -- & -- & -- \\
After BoolQ & 80.4$\pm$0.8 & 71.2$\pm$1.7 & 48.4$\pm$2.2 & 82.2$\pm$1.6 & 88.0$\pm$1.3 & 91.0$\pm$0.5 & 95.2$\pm$0.3 & 79.3$\pm$0.8 & 85.3$\pm$0.3 & -- & -- \\
After HotpotQA & 78.7$\pm$2.3 & 73.7$\pm$1.7 & 47.7$\pm$0.9 & 80.7$\pm$1.6 & 88.3$\pm$0.8 & 91.3$\pm$1.0 & 94.7$\pm$0.8 & 79.0$\pm$0.9 & 85.7$\pm$0.8 & 69.9$\pm$1.8 & -- \\
After DROP (all) & 79.2$\pm$0.8 & 73.4$\pm$1.4 & 48.1$\pm$2.5 & 80.5$\pm$0.5 & 87.8$\pm$0.3 & 90.8$\pm$1.2 & 95.2$\pm$0.8 & 79.7$\pm$1.0 & 85.8$\pm$1.2 & 67.4$\pm$0.5 & 70.7$\pm$0.5 \\
\midrule
Single-skill & 81.2$\pm$1.5 & 76.9$\pm$1.7 & 41.3$\pm$2.8 & 80.0$\pm$1.5 & 87.5$\pm$0.9 & 91.5$\pm$0.7 & 95.0$\pm$0.6 & 78.0$\pm$0.9 & 87.0$\pm$0.8 & 67.4$\pm$1.3 & 70.3$\pm$0.5 \\
\bottomrule
\end{tabular}
\caption{Sequential registration of eleven skills into one Qwen3.5-4B
interface (mean $\pm$ standard deviation over three registration seeds). The
order follows the rows. A dash means the skill is not yet registered. Single-skill registers only that skill. }
\label{tab:app-retention-ext}
\end{table*}

\paragraph{RL-trained steering baselines.}
\label{sec:app-steering}

Table~\ref{tab:app-steering} compares the reward-learned KV-Skill with
RL-trained bias-only steering, which adds a learned vector with a fixed
coefficient at each layer. Both use the reward protocol of
Appendix~\ref{sec:reward-details}. Bias-only steering at the four injection
depths approaches KV-Skill on SearchQA but trails it on LiveMath; extending it
to all 32 layers does not close the gap.

\begin{table}[H]
\centering
\small
\setlength{\tabcolsep}{5pt}
\renewcommand{\arraystretch}{1.08}
\begin{tabular}{@{}lcc@{}}
\toprule
\textbf{Qwen3.5-4B method} & \textbf{SearchQA} & \textbf{LiveMath} \\
\midrule
KV-Skill               & 80.0$\pm$3.4 & 75.6$\pm$4.1 \\
Bias-only, 4 depths    & 78.9$\pm$0.9 & 69.3$\pm$4.1 \\
Bias-only, all layers  & 77.2$\pm$0.2 & 66.9$\pm$5.7 \\
\bottomrule
\end{tabular}
\caption{RL-trained bias-only steering versus the reward-learned KV-Skill on
Qwen3.5-4B (SearchQA EM, LiveMath accuracy; mean $\pm$ standard deviation over
seeds). Values match the corresponding rows of Table~\ref{tab:substrate}.}
\label{tab:app-steering}
\end{table}

\subsection{Sequential Registration and Replay}
\label{sec:app-retention}

We register SearchQA, LiveMath, and STaRK-Prime in this order into one
Qwen3.5-4B interface.

We run the replay ablation on SearchQA with seed 0. Without replay, performance
falls from 78.7 after the first stage to 76.3 after the third. A matched
self-replay control reaches 78.3, while mixed replay of earlier skills reaches
80.3. Thus, additional optimization is sufficient to prevent forgetting in
this sequence, while replaying prior skills supplies the further improvement.

We extend this sequence to eleven reported skills under the same protocol
(Table~\ref{tab:app-retention-ext}). Each row is the shared interface after the
named skill is registered, and each column reads one skill through its own
operator. After all eleven stages, the shared interface averages 78.1 across the eleven
skills, against 77.8 when each skill has its own interface (the Single-skill row).
Seven skills match or beat their own-interface accuracy, and nine end within
two points. LiveMath drops the most (76.9 to 73.4). STaRK-Prime instead gains
6.8 points (41.3 to 48.1). A possible reason is that registering more skills
reduces overfitting of the shared interface. Table~\ref{tab:stark} points the
same way: STaRK-Prime registered after nine other skills reaches 49.7. Replay keeps training earlier skills, so they can improve in later
stages. LiveMath, for example, rises from 65.1 after its own stage to 75.8
after STaRK-Prime.

\subsection{Long-Horizon Agentic Tasks}
\label{sec:app-agentic}

On ALFWorld, reward learning lifts Base from 30.6 to 42.3. Behavior cloning
from successful trajectories reaches 69.4 with the same frozen backbone, and
SkillOpt reaches 81.3. This contrast
supports credit assignment, rather than operator capacity, as the main
bottleneck in this setting.

\begin{table}[H]
\centering
\scriptsize
\setlength{\tabcolsep}{3pt}
\renewcommand{\arraystretch}{1.08}
\begin{tabular}{@{}lccc@{}}
\toprule
\textbf{Method}
& \shortstack{\textbf{OfficeQA}\\{\scriptsize Success}}
& \shortstack{\textbf{SpreadsheetBench}\\{\scriptsize Success}}
& \shortstack{\textbf{ALFWorld}\\{\scriptsize Success}} \\
\midrule
Base                     & 14.5 & 9.3  & 30.6 \\
Text Skill               & 29.7 & --   & -- \\
SkillOpt                 & 29.7 & 23.9 & 81.3 \\
KV-Skill (RL scratch)    & 37.2 & 22.3 & 40.3 \\
KV-Skill (RL from skill) & \textbf{37.8} & \textbf{25.8} & 42.3 \\
\bottomrule
\end{tabular}
\caption{Long-horizon task success on Qwen3.5-4B. ALFWorld exposes the current
limitation of learning from sparse terminal reward. Text Skill is reported on OfficeQA only.}
\label{tab:app-agentic}
\end{table}

\section{Resource Accounting}
\label{sec:efficiency}

KV-Skill changes where task knowledge is stored and how it is accessed. It does
not provide a uniform efficiency advantage. Text skills are smallest on disk,
but their tokens are processed on every query. KV-Skills add no prompt
positions, but require an external operator and a shared per-backbone
interface. We quantify these costs using measured file sizes, parameter counts,
token counts, and latency.

Unless otherwise stated, token counts and latency measurements use
Qwen3.5-4B. File sizes are measured with \texttt{os.path.getsize}, and
parameter counts are obtained by summing \texttt{numel} over the loaded state
dictionary. Latency is measured on one H100 with a 512-slot KV-Skill.

\begin{table*}[t]
\centering
\small
\setlength{\tabcolsep}{5pt}
\renewcommand{\arraystretch}{1.08}
\begin{tabular}{@{}lrrc@{}}
\toprule
\textbf{Method}
& \textbf{Skill-side parameters}
& \textbf{Per-skill artifact}
& \textbf{Added prompt positions} \\
\midrule
Text Skill
& 0
& 2.7--13 KiB
& 671--2806 \\
Soft prefix
& 81,920
& 165,509 B
& 32 \\
LoRA ($r=8$)
& 5,242,880
& 10.0 MiB FP16
& 0 \\
Compact reward-learned KV-Skill
& 16,384
& 67,229 B
& 0 \\
Text-derived KV-Skill, current FP32
& 0 trainable
& 40.0 MiB
& 0 \\
Text-derived KV-Skill, tied FP16
& 0 trainable
& 10.0 MiB
& 0 \\
Text-derived KV-Skill, rank one
& 0 trainable
& 20 KiB
& 0 \\
\bottomrule
\end{tabular}
\caption{\textbf{Per-skill resource comparison.}
The compact KV-Skill contains two $32\times256$ FP32 factors. The soft prefix
contains 32 BF16 vectors of dimension 2560. The LoRA size is computed
analytically from the rank-8 configuration used by SoftSkill and was not
reproduced in our implementation. KV-Skill sizes exclude the shared
per-backbone interface $I$.}
\label{tab:per-skill-storage}
\end{table*}

\begin{table*}[t]
\centering
\small
\setlength{\tabcolsep}{5pt}
\renewcommand{\arraystretch}{1.08}
\begin{tabular}{@{}lrrr@{}}
\toprule
\textbf{Interface}
& \textbf{File size}
& \textbf{Interface parameters}
& \textbf{Total trainable parameters} \\
\midrule
Qwen3.5-4B, universal
& 53,169,167 B
& 13,109,768
& 13,283,336 \\
Qwen3.5-4B, registered on nine skills
& 53,169,991 B
& 13,109,768
& 13,283,336 \\
gpt-oss-20b, universal
& 66,949,221 B
& 16,591,688
& 16,732,808 \\
Compact interface, $d_s=256$
& 5,983,247 B
& 1,313,288
& 1,486,856 \\
\bottomrule
\end{tabular}
\caption{\textbf{Shared interface cost.}
Total trainable parameters include the interface and the trained RMSNorm
scales. The interface is stored once per backbone rather than once per
KV-Skill.}
\label{tab:interface-cost}
\end{table*}

\subsection{Text Skills Are Small but Recur on Every Query}

Text skills require almost no artifact storage. The four skills in
Table~\ref{tab:text-skill-cost} occupy only 2.7--13 KiB. Their cost instead
appears in the prompt. Each call processes an additional 671--2806 tokens,
depending on the skill.

\begin{table}[H]
\centering
\small
\setlength{\tabcolsep}{5pt}
\renewcommand{\arraystretch}{1.08}
\begin{tabular}{@{}lrr@{}}
\toprule
\textbf{Text skill} & \textbf{File size} & \textbf{Tokens per query} \\
\midrule
LiveMath       & 3,343 B  & 671  \\
SearchQA       & 9,941 B  & 2,035 \\
STaRK-Prime    & 2,767 B  & 785  \\
ALFWorld       & 13,179 B & 2,806 \\
\bottomrule
\end{tabular}
\caption{\textbf{Storage and recurring prompt cost of text skills.}
The file is stored once, but all skill tokens are added to every query.}
\label{tab:text-skill-cost}
\end{table}

This difference is visible in our measured prompts. On LiveMath, SearchQA, and STaRK-Prime, Text Skill therefore adds 671, 2035,
and 785 tokens to every query, whereas KV-Skill leaves the prompt unchanged
because its operator does not enter the attention sequence.

The statement that KV-Skill does not grow with the conversation applies to the
external operator. The model's ordinary attention KV cache still grows with
the interaction history.

\subsection{Per-Skill Storage}

A text-derived KV-Skill contains one factor pair at each of $G_s$ injection
layers. With $N_s$ slots, skill dimension $d_s$, factor multiplicity $f$, and
$b$ bytes per scalar, its storage is
\begin{equation}
\mathrm{Storage}(M_s)
=
fG_sN_sd_sb.
\label{eq:derived-storage}
\end{equation}
The current implementation stores separate FP32 factors, even though they are
bit-identical in the text-derived construction. It therefore uses $f=2$ and
$b=4$. Tying the factors gives $f=1$, and FP16 storage gives $b=2$.

For $G_s=4$, $N_s=512$, and $d_s=2560$, the current representation occupies
40 MiB. Tying the factors and using FP16 reduces this to 10 MiB. The current
production evaluation does not persist this artifact; it derives the operator
from its text skill when the evaluation begins. The same tensor storage is
nevertheless required while the operator is loaded.

The rank-one representation requires one tied FP16 direction per injection
layer:
\begin{equation}
4 \times 2560 \times 2
=
20{,}480 \ \text{bytes},
\end{equation}
or approximately 20 KiB. This is the marginal size of the KV-Skill operator,
not the total deployment size.

The comparison exposes two different operating points. The full 512-slot
text-derived operator is four times larger than the rank-8 LoRA configuration
as currently stored, and equal in size after factor tying and FP16 conversion.
The rank-one operator is only 20 KiB, which is $500\times$ smaller than this
LoRA configuration. The compact reward-learned KV-Skill occupies approximately
64 KiB of tensor payload.

These numbers should not be interpreted as a universal storage advantage.
Text skills remain much smaller as files. KV-Skill instead avoids paying their
token cost on every query.

\subsection{The Shared Interface}

KV-Skill also requires a model-specific interface $I$. This is not a free
component. It is trained once for each backbone and shared across all
registered KV-Skills for that backbone.

For Qwen3.5-4B, the same registered interface serves nine KV-Skills without
increasing its parameter count. Its fixed cost is therefore amortized as the
number of registered KV-Skills grows. Comparisons based only on the 20 KiB or
64 KiB operator should be understood as marginal per-skill comparisons after
this shared interface has been installed.

\subsection{Compute to First Token}
\label{sec:app-flops}
Reading a KV-Skill computes $W_s(U_s^{\top}q)$ without forming $M_s$, so its
cost per token is linear in the number of slots and has no $N_s^2$ term.
Table~\ref{tab:app-flops} reports analytic FLOPs to the first token at a
query length of $T=512$, relative to Base, together with the extra FLOPs per
generated token. Once the text skill's prefix is cached, its time-to-first-token
cost is close to that of the registered operator; the remaining differences are
that KV-Skill occupies no prompt positions and that its per-token cost after
compression does not depend on skill length.

\begin{table}[H]
\centering
\scriptsize
\setlength{\tabcolsep}{3pt}
\renewcommand{\arraystretch}{1.08}
\begin{tabular}{@{}p{0.2\columnwidth}ccccc@{}}
\toprule
$T=512$ & Text unc. & Cached & Full & RL-KV & Rank-1 \\
\midrule
LiveMath  & 2.32$\times$ & 1.012$\times$ & 1.018$\times$ & 1.0015$\times$ & 1.00001$\times$ \\
SearchQA & 5.07$\times$ & 1.037$\times$ & 1.018$\times$ & 1.0015$\times$ & 1.00001$\times$ \\
ALFWorld & 6.65$\times$ & 1.051$\times$ & 1.018$\times$ & 1.0015$\times$ & 1.00001$\times$ \\
\midrule
Extra FLOPs per generated token & 267M & 267M & 126M & 10.6M & 41K \\
\bottomrule
\end{tabular}
\caption{Analytic FLOPs to the first token on Qwen3.5-4B, relative to Base, at
query length $T=512$. Rows give the LiveMath, SearchQA, and ALFWorld text
skills (671, 2,035, and 2,806 tokens); the last row gives the extra FLOPs per
generated token for the SearchQA skill. Text unc.\ and Cached are
the text skill without and with prefix caching; Full is the full-rank registered
512-slot operator; RL-KV is the compact reward-learned operator; Rank-1 is the
rank-one registered operator.}
\label{tab:app-flops}
\end{table}

\subsection{Latency and Context-Length Scaling}

We isolate the KV-Skill branch by comparing the injector-on and injector-off
conditions at matched token counts. Prefill overhead is 1.8\% at 128 tokens
and approximately 1.0\% at 512 and 2048 tokens. Generating a fixed sequence of
64 new tokens introduces a 6.5\% decoding overhead.

\begin{table}[H]
\centering
\small
\setlength{\tabcolsep}{5pt}
\renewcommand{\arraystretch}{1.08}
\begin{tabular}{@{}lc@{}}
\toprule
\textbf{Measurement} & \textbf{Result} \\
\midrule
Prefill overhead, $T=128$  & 1.8\% \\
Prefill overhead, $T=512$  & 1.0\% \\
Prefill overhead, $T=2048$ & 1.0\% \\
Decode overhead, 64 tokens & 6.5\% \\
\midrule
Growth overhead, $T=128$   & 6.6\% \\
Growth overhead, $T=512$   & 4.7\% \\
Growth overhead, $T=2048$  & 6.5\% \\
Growth overhead, $T=8192$  & 5.0\% \\
Peak-memory growth attributable to context
& $<0.001$ GiB \\
\bottomrule
\end{tabular}
\caption{\textbf{Measured KV-Skill runtime overhead on one H100.}
The experiment uses Qwen3.5-4B and a 512-slot KV-Skill. Peak-memory differences
round to 0.000 GiB at every evaluated context length.}
\label{tab:latency}
\end{table}

The overhead remains between 4.7\% and 6.6\% as context length grows from 128
to 8192 tokens. The measured context-dependent peak-memory difference is below
the reporting resolution at every length. These results support the structural
claim that the operator does not enter the attention KV cache. They do not
imply zero memory use: the operator and interface still occupy a fixed amount
of memory.

KV-Skill is therefore not cheaper at a matched token count. Its end-to-end
advantage over prompt-based skills comes from processing fewer input tokens
and, in many tasks, producing shorter answers. Shorter answers reflect task behavior, not lower per-token computation.

\subsection{Loading and Swapping}

A cached KV-Skill is selected through tensor assignment. The measured
assignment time is below timer resolution and is reported as 0.00 ms. Deriving
an uncached 512-token operator takes 158 ms. Applying the optional rank-one SVD
takes an additional 432 ms and can be performed offline. By comparison,
prefilling a 512-token text skill takes 155 ms and leaves those 512 positions
in the context.

\begin{table}[H]
\centering
\small
\setlength{\tabcolsep}{5pt}
\renewcommand{\arraystretch}{1.08}
\begin{tabular}{@{}lc@{}}
\toprule
\textbf{Operation} & \textbf{Measured time} \\
\midrule
Select cached KV-Skill        & 0.00 ms \\
Derive uncached 512-token operator & 158 ms \\
Rank-one SVD                  & 432 ms \\
Prefill 512-token text skill  & 155 ms \\
\bottomrule
\end{tabular}
\caption{\textbf{Loading and switching costs.}
Derivation and SVD are one-time operations and can be performed before
deployment. Text-skill prefill is repeated whenever the text skill must be
inserted into a new context.}
\label{tab:swap-cost}
\end{table}

The cached-swap measurement covers changing the operator used for future
tokens. It does not recompute representations already stored in the model's
attention KV cache. If an application requires the entire history to be
reinterpreted under the newly selected KV-Skill, that history must be
prefilled again.

\subsection{Summary}

The measurements reveal a three-way tradeoff. Text skills minimize artifact
storage but consume prompt positions on every query. Full text-derived
KV-Skills preserve the context window but can require more storage than LoRA.
Rank reduction makes their marginal per-skill storage very small, while the
shared interface remains a substantial one-time cost. Finally, the operator
read adds a modest and approximately context-independent runtime overhead.
KV-Skill should therefore be understood as a context-preserving and modular
representation, not as uniformly cheaper inference.

\end{document}